# A CONCATENATING FRAMEWORK OF SHORTCUT CONVOLUTIONAL NEURAL NETWORKS

Yujian Li (liyujian@bjut.edu.cn), Ting Zhang, Zhaoying Liu, Haihe Hu

ABSTRACT

It is well accepted that convolutional neural networks play an important role in learning excellent features for image classification and recognition. However, in tradition they only allow adjacent layers connected, limiting integration of multi-scale information. To further improve their performance, we present a concatenating framework of shortcut convolutional neural networks. This framework can concatenate multi-scale features by shortcut connections to the fully-connected layer that is directly fed to the output layer. We do a large number of experiments to investigate performance of the shortcut convolutional neural networks on many benchmark visual datasets for different tasks. The datasets include AR, FERET, FaceScrub, CelebA for gender classification, CUReT for texture classification, MNIST for digit recognition, and CIFAR-10 for object recognition. Experimental results show that the shortcut convolutional neural networks can achieve better results than the traditional ones on these tasks, with more stability in different settings of pooling schemes, activation functions, optimizations, initializations, kernel numbers and kernel sizes.

## 1 INTRODUCTION

Convolutional neural networks (CNNs) are hierarchical feed-forward architectures that compute progressively in invariant representations of the input images. As an excellent method for extracting image features, they have been widely applied to a variety of domains, such as face recognition (Taigman et al, 2014) (Lopes et al, 2017) (Sun et al, 2016), bounding box object detection (Girshick et al, 2014) (Zhang et al, 2016), key point prediction (Sun et al, 2013) (Jonathan et al, 2014), and large-scale image classification task (Simonyan and Zisserman, 2014) (Deng et al, 2009) (Szegedy et al, 2015) (Fukushima. 1979), etc.

The first implemented CNN is considered to be the model of neocognitron developed by Fukushima with the insight of receptive field (Fukushima. 1979). In 1998, LeCun *et al*. combined convolutional layers with pooling layers to make an early version of modern CNNs (LeNet) (LeCun et al, 1998). In 2012, Krizhevsky et al. (2012) proposed a breakthrough architecture, the AlexNet, for ImageNet Large Scale Visual Recognition Competition (ILSVRC). In 2013, Simonyan and Zisserman presented the VGG network, using very small convolution filters to push the depth of weight layers (Simonyan and Zisserman, 2014). In 2014, by integrating with "inception modules", Szegedy et al. designed the GoogLeNet (Szegedy et al, 2015). It is worth mentioning that the AlexNet, the VGG network, and the GoogLeNet won the first place in ILSVRC 2012, 2013, and 2014, respectively.

Traditionally, a standard CNN is composed of convolutional layers (CLs), pooling layers (PLs),

and fully-connected layers (FCLs), as illustrated in Fig.1. It can be seen that CLs and PLs are generally arranged in an alternating fashion to extract features from different scales. One or more FCLs together with the output layer are exploited to work as a classifier. For convenience, we refer to a CL or a PL as a CPL. In fact, a CNN comprises a number of CPLs, followed by several FCLs. Note that a CL/PL may consist of many convolutional/pooling feature maps. One major advantage of CNNs is the use of shared weights in CLs, which means that the same convolutional kernel is used for each pixel in the layer. This not only greatly reduces the number of parameters involved in a CNN, but also improves its performance (Jin et al, 2016).

Strictly speaking, a standard CNN has no shortcut connections cross layers, where the topmost CPL separates the lower CPLs from the first fully-connected layer (FFCL). Thus, given the features extracted from the topmost CPL, the final output of the FCLs is independent of the lower CPLs. Although such a CNN can work very well in many situations, it is limited to integrate multi-scale information from an image.

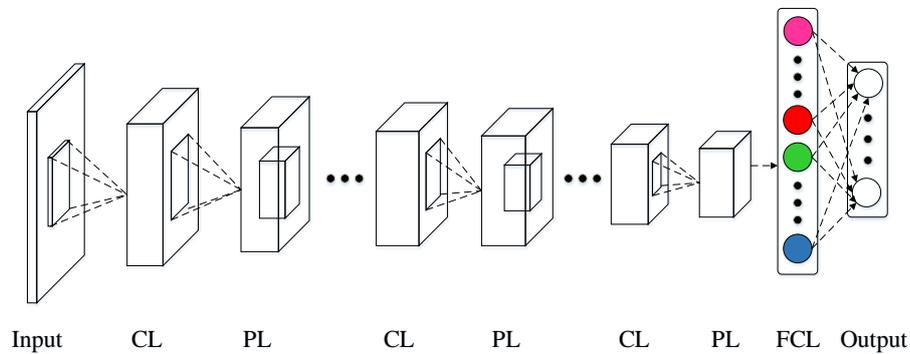

Fig.1 An example of standard CNNs. There is only one FCL in this architecture.

To make use of discriminative information from non-topmost CPLs, we propose a concatenating framework of shortcut convolutional neural networks (S-CNNs), to integrate multi-scale features through shortcut connections in a CNN. This framework can select some different levels of powerful features to be concatenated for final decision of classification and recognition. As displayed in Fig. 2, an S-CNN can integrate multi-scale features through shortcut connections from a number of CPLs to the FFCL. It is well admitted that human vision is a multi-scale process (Donoho and Huo, 2001). Therefore, it would be reasonable to integrate different levels of image features for robust classification, where low-level features are finer and high-level features are more invariant, with their combination probably producing good representations in leverage of concrete and abstraction (Sermanet et al, 2013).

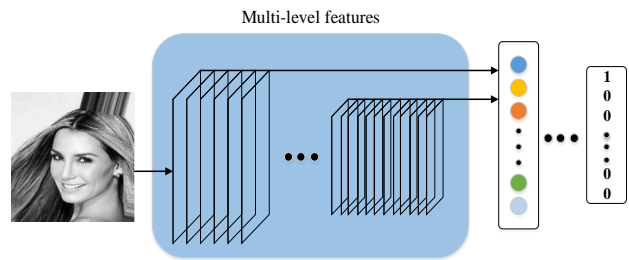

Fig. 2. An architecture of S-CNNs, where some different CPLs are concatenated to form the FFCL through shortcut connections.

Recently, shortcut connections have attracted great interest since the success of residual networks on the tasks of ILSVRC 2015 classification, ImageNet detection, ImageNet localization, COCO detection, and COCO segmentation (He et al, 2016). The related work of S-CNNs could be roughly divided into two modes: trainable and fixed.

Trainable-mode S-CNNs refer to the CNNs that have trainable shortcut connections. For example, Sermanet and LeCun applied a multi-scale CNN to the task of traffic sign classification (Sermanet and LeCun, 2011), and they got the first place in the German Traffic Sign Recognition Benchmark (GTSRB) competition. In their network, both the first pooling layer and the second pooling layer are directly fed to the fully-connected layer through trainable shortcut connections. Sun et al. proposed a DeepId network for face verification (Sun et al, 2014), which only allows the last CPL but one to have trainable shortcut connections. Srivastava et al. (2015) designed a highway network for digit classification, allowing earlier representations to flow unimpededly to later layers through parameterized shortcut connections known as "information highways". The parameters of shortcut connections are learned for controlling the amount of information allowed on these "highways".

The fixed mode S-CNNs refer to the CNNs that have fixed shortcut connections. For example, the deep residual networks allow shortcut connections to cross two or three convolutional layers (He et al, 2016). Huang et al. presented densely connected CNNs (DenseNets), which allow connections from each layer to every other layer in a feed-forward fashion (Huang et al, 2017). Vincent et al. proposed a texture and shape CNN for texture classification, which only allows the shortcut connections to cross three CPLs (Andrearczyk and Whelan, 2016). Shen et al. presented a multi-crop CNN for lung nodule malignancy suspiciousness classification, which concatenates three multi-crop pooling layers (Shen et al, 2017). Liu et al. introduced a single shot multibox detector for detecting objects in images using a single deep neural network, which concatenates multiple convolutional features (Liu et al, 2016). These shortcut connections are always alive and the gradients can easily back propagate through them, which results in faster training.

Currently, both the trainable-mode and fixed-mode S-CNNs are discussed as a specific structure. Apart from them, we present a concatenating framework of multi-scales features instead of a specific structure. In contrast to the trainable-mode S-CNNs, our framework has the fixed value of 1 for all weights of shortcut connections. And compared with the fixed-mode CNNs, our framework can bypass more than three hidden layers. Overall, the motivation of our work is to integrate multi-scale features through a variety of shortcut connections with fixed weights.

In this paper, we propose a concatenating framework of S-CNNs by adding shortcut connections to standard CNNs, together with a shortcut backpropagation algorithm. Using an indicator of binary string (called shortcut indicator), we can conveniently choose a shortcut style from the framework to integrate different levels of multi-scale features. Based on this convenience, we conduct a large number of experiments to compare S-CNNs with standard CNNs on seven datasets for classification of gender and texture as well as for recognition of digit and object. Moreover, we compare their performance in different settings of pooling schemes, activation functions, initializations, optimizations, and convolutional kernels' numbers and sizes. Additionally, experimental results show that S-CNNs can generally achieve better performance than standard CNNs with more stability. Finally, we summarize the whole paper in conclusions.

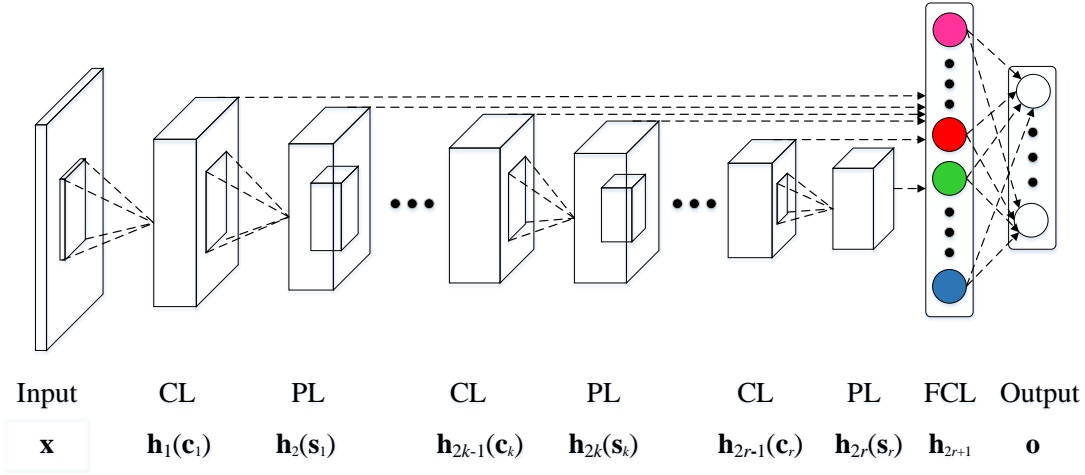

Fig. 3. A concatenating framework of S-CNNs. In this framework, features from different CPLs are concatenated to form the FCL which is directly fed to the output layer. Here, the shortcut indicator is 111⋯111, indicating the shortcut style of all lower 2r-1 CPLs having shortcut connections to the FCL.

## 2 FRAMEWORK DESCRIPTION

Based on the architecture of CNNs in Fig.1, we present a concatenating framework of S-CNNs by adding shortcut connections. As displayed in Fig. 3, this framework is an alternating structure of $r$ CLs and $r$ PLs, followed by a FCL and an output layer. The FCL is a concatenation of these CLs and PLs through a style of shortcut connections, which is represented by a binary string, called shortcut indicator (SI). Accordingly, we can give a description of the framework as follows.

The input $\mathbf{x}$ is a 3-dimensional array of size $h \times w \times n$, where $h$ and $w$ are spatial dimensions, and $n$ is the channel dimension, with $n=3$ for color images and $n=1$ for grayscale images.

Using "$*$" to stand for convolutional operator and "$f$" for activation function, the computation of a convolutional layer can be expressed as

$$\mathbf{h}^l_{2k-1,j} = \mathbf{c}^l_{k,j} = f\left(\mathbf{u}^l_{2k-1,j}\right) = f\left(\sum_i \mathbf{h}^l_{2k-2,i} * \mathbf{W}^{2k-1}_{ij} + \mathbf{b}^{2k-1}_j\right), 1 \leq k \leq r, \quad (1)$$

where $\mathbf{W}^{2k-1}_{ij}$ is the weight matrix between the $i$-th feature map in the (2k-2)-th hidden layer and the $j$-th feature map in the (2k-1)-th hidden layer, $\mathbf{b}^{2k-1}_j$ is the bias of the $j$-th feature map in the (2k-1)-th hidden layer. $\mathbf{c}^l_{k,j}$ stands for the $j$-th feature map in the $k$-th CL, with $\mathbf{h}^l_{2k-2,i}$ and $\mathbf{h}^l_{2k-1,j}$ denoting the $i$-th feature map in the (2k-2)-th hidden layer and the $j$-th feature map in the (2k-1)-th hidden layer for the $l$-th sample, respectively. $f$ can be sigmoid (Ni et al, 2013) or rectified linear unit (ReLU) (Krizhevsky, et al, 2012). Here, we let $\mathbf{h}^l_0 = \mathbf{x}^l$.

In each pooling layer, we use a fixed stride for all feature maps. The pooling function is formulated as:

$$\mathbf{h}^l_{2k,j} = \mathbf{s}^l_{k,j} = pooling\{\mathbf{h}^l_{2k-1,j}\}, 1 \leq k \leq r, \quad (2)$$

where $pooling\{\cdot\}$ can be average pooling or max-pooling. $\mathbf{h}^l_{2k-1,j}$ and $\mathbf{h}^l_{2k,j}$ indicate the $j$-th feature map in the (2k-1)-th hidden layer and the $j$-th feature map in the 2k-th hidden layer for the $l$-th sample, respectively. $\mathbf{s}^l_{k,j}$ stands for the $j$-th feature map in the $k$-th PL.

The FCL is the concatenation of two or more CPL activations through shortcut connections, forming the entire discriminative vector of multi-scale features. In fact, the FCL takes the form:

$$\mathbf{h}^l_{2r+1} = \left(a_1\mathbf{h}^l_1, a_2\mathbf{h}^l_2, \cdots, a_{2k-1}\mathbf{h}^l_{2k-1}, a_{2k}\mathbf{h}^l_{2k}, \cdots, \mathbf{h}^l_{2r}\right), \tag{3}$$

where $\mathbf{h}^l_{2k-1}$ and $\mathbf{h}^l_{2k}$ ($1 \leq k \leq r$) denote the (2k-1)-th and the 2k-th hidden layer for the *l*-th sample, respectively. Let $A = a_1 a_2 a_3 \cdots a_{2r-1}$ be a binary string called shortcut indicator, which indicates the shortcut style. For example, $A = 111\cdots1$ indicates the shortcut style that all the 2r-1 associated CPLs have shortcut connections to the FCL. $A = 100\cdots0$ represents the shortcut style that only the first CPL has shortcut connections to the FCL. $A = 000\cdots0$ denotes the shortcut style that has no shortcut connections at all, meaning the standard CNN.

The actual output is a *C*-way softmax predicting the probability distribution over *C* different classes, expressed as:

$$\mathbf{o}^l = softmax(\mathbf{u}^l) = softmax\left(\mathbf{W}^{2r+2}\mathbf{h}^l_{2r+1} + \mathbf{b}^{2r+2}\right), \tag{4}$$

where $\mathbf{W}^{2r+2}$ and $\mathbf{b}^{2r+2}$ stand for the weight and bias of the output layer, with $softmax_i(\mathbf{x}) = \exp(x_i) / \sum_j \exp(x_j)$.

## 3 LEARNING ALGORITHM

For the *l*-th sample, the S-CNNs compute the activations of all CPLs, the FCL and the actual output as follows:

$$\begin{cases} \mathbf{h}^l_{2k-1,j} = f\left(\mathbf{u}^l_{2k-1,j}\right) = f\left(\sum_i \mathbf{h}^l_{2k-2,i} * \mathbf{W}^{2k-1}_{ij} + \mathbf{b}^{2k-1}_j\right), 1 \leq k \leq r \\ \mathbf{h}^l_{2k,j} = pooling\left\{\mathbf{h}^l_{2k-1,j}\right\}, 1 \leq k \leq r \\ \mathbf{h}^l_{2r+1} = \left(a_1\mathbf{h}^l_1, a_2\mathbf{h}^l_2, \cdots, a_{2k-1}\mathbf{h}^l_{2k-1}, a_{2k}\mathbf{h}^l_{2k}, \cdots, \mathbf{h}^l_{2r}\right) \\ \mathbf{o}^l = softmax(\mathbf{u}^l) = softmax\left(\mathbf{W}^{2r+2}\mathbf{h}^l_{2r+1} + \mathbf{b}^{2r+2}\right) \end{cases}, \tag{5}$$

Let $\mathbf{y}^l = \left(y^l_1, y^l_2, \ldots, y^l_C\right)^T$ be the desired output and $\mathbf{o}^l = \left(o^l_1, o^l_2, \ldots, o^l_C\right)^T$ the actual output. Taking the objective function of cross entropy loss, namely,

$$L_N\left(\mathbf{y}^l, \mathbf{o}^l\right) = -\sum_{l=1}^N \sum_{c=1}^C y^l_c \log\left(o^l_c\right), \tag{6}$$

we can first compute the sensitivities $\boldsymbol{\delta}^l_k$ ($1 \leq k \leq 2r+1$) of each hidden layer and the sensitivity $\boldsymbol{\delta}^l$ of the output layer as follows:

$$\begin{cases} \boldsymbol{\delta}^l = \mathbf{o}^l - \mathbf{y}^l \\ \boldsymbol{\delta}^l_{2r+1} = \left[\left(\mathbf{W}^{2r+2}\right)^T \boldsymbol{\delta}^l\right] \circ softmax'\left(\mathbf{u}^l_{2r+2}\right) \\ \left(\boldsymbol{\delta}^l_{1,FC}\ \boldsymbol{\delta}^l_{2,FC}\ \cdots \boldsymbol{\delta}^l_{2k-1,FC}\ \boldsymbol{\delta}^l_{2k,FC}\ \cdots \boldsymbol{\delta}^l_{2r,FC}\right) = \boldsymbol{\delta}^l_{2r+1} \\ \boldsymbol{\delta}^l_{2r} = \boldsymbol{\delta}^l_{2r,FC} \\ \boldsymbol{\delta}^l_{2k-1,j} = f'\left(\mathbf{u}^l_{k,j}\right) \circ uppooling\left\{\boldsymbol{\delta}^l_{2k,j}\right\} + a_{2k-1}\boldsymbol{\delta}^l_{2k-1,FC,j}, 1 \leq k \leq r \\ \boldsymbol{\delta}^l_{2k,j} = \boldsymbol{\delta}^l_{2k+1,j} * rot180\left(\mathbf{W}^{2k+1}_{ij}\right) + a_{2k}\boldsymbol{\delta}^l_{2k,FC,j}, 1 \leq k \leq r \end{cases}, \tag{7}$$

where $\boldsymbol{\delta}^l$ stands for the sensitivity (or backpropagation error) of the output layer, $\boldsymbol{\delta}^l_{2k-1}$ ($1 \leq k \leq r$) and $\boldsymbol{\delta}^l_{2k}$ ($1 \leq k \leq r$) represent the sensitivities of the (2k-1)-th and 2k-th hidden layer, respectively. $\boldsymbol{\delta}^l_{2k-1,FC}$ or $\boldsymbol{\delta}^l_{2k,FC}$ ($1 \leq k \leq r$) is the part of the (2r+1)-th hidden layer (i.e. the FCL) that corresponds to the (2k-1)-th or 2k-th hidden layer. Additionally, $uppooling\{\cdot\}$ is the upsampling function for the pooling function defined by (2). $softmax'(\cdot)$ stands for the derivative of the softmax function, $rot180(\cdot)$ indicates flipping a matrix horizontally and vertically, and the symbol "∘" denotes Hadamard product.

Using (6) and (7), we can compute the derivatives with respect to each parameter (i.e., weights and biases) as follows.

$$\begin{cases} \dfrac{\partial L_N}{\partial \mathbf{W}^{2r+2}} = \sum_{l=1}^{N} \boldsymbol{\delta}^l \left(\mathbf{h}_{2r+1}^l\right)^T, \dfrac{\partial L_N}{\partial \mathbf{b}^{2r+2}} = \sum_{l=1}^{N} \boldsymbol{\delta}^l, \\ \dfrac{\partial L_N}{\partial \mathbf{W}_{ij}^{2k-1}} = \sum_{l=1}^{N} \boldsymbol{\delta}_{2k-1,j}^l * \mathbf{h}_{2k-2,i}^l, \dfrac{\partial L_N}{\partial \mathbf{b}_j^{2k-1}} = \sum_{l=1}^{N} \boldsymbol{\delta}_{2k-1,j}^l, 1 \le k \le r \end{cases} \quad (8)$$

Based on (5)-(8), we design a training algorithm of gradient descent for the S-CNNs as shown in **Algorithm 1**, i.e. shortcut BP for S-CNNs. Note that *maxepoch* stands for the number of training iterations.

**Input**: Training set $S = \{(\mathbf{x}^l, \mathbf{y}^l), 1 \le l \le N\}$, network architecture, *maxepoch*
**Output**: network parametets
    Randomly initialize the weights and biases of the S-CNNs;
    **for** *epoch*=1 to *maxepoch* do
     **for** *l*=1 to *N* do
      Compute the hidden activations and the actual outputs by (5);
      Compute the sensitivities of each layer by (7);
      Compute the derivatives by (8);
      Update all the weights and biases with gradient descent;
     **end**
    **end**

**Algorithm 1:** Shortcut BP for S-CNNs

## 4 EXPERIMENTS

In this section, we evaluate S-CNNs for gender classification, texture classification, digit recognition, and object recognition. We implemented a stochastic version of Algorithm 1 by the GPU-accelerated ConvNet library Caffe (Jia. 2013), initializing the weights by the "Xavier" method (Glorot and Bengio, 2010) to train the S-CNNs. The experimental environment is a desktop PC equipped with E5-2643 V3 CPU, 64GB memory and a NVIDIA Tesla K40c.

In all experiments, the momentum is set to 0.9 and the mini-batch size is set to 100. The weight decay is set to 0.004 for gender classification, texture classification and object recognition, and to 0.005 for digit recognition. The fixed learning rate is set to 0.001 for weights and double for biases in all the four tasks.

### 4.1 GENDER CLASSIFICATION

In this subsection, we use four datasets, namely, AR (Maetinez. 2001), FERET (Phillips, et al, 1998), FaceScrub (Ng and Winkle, 2014), and CelebA (Yang etal, 2015), to compare the performance of the standard CNN and S-CNNs on gender classification. The training iterations are 5000, 5000, 50000, and 60000 for them, respectively. With their examples shown in Fig.4, we describe some more detailed information as follows.

(1) The AR dataset consists of over 4000 frontal images for 126 subjects, including different

facial expressions, illumination conditions and disguises. Only a subset of 50 male subjects and 50 female subjects were used in the experiments, 26 images per subject. From the subset, 40 females and 40 males were selected for training, and the rest for testing.

(2) The FERET dataset contains 14038 images of 1196 different individuals with at least 5 images each, including different lighting conditions and non-neural expressions. In the dataset, there are 5195 female images and 8843 male images, respectively. All the images were used in the experiments, where 4414 female images and 4417 male images were selected for training, and the rest for testing.

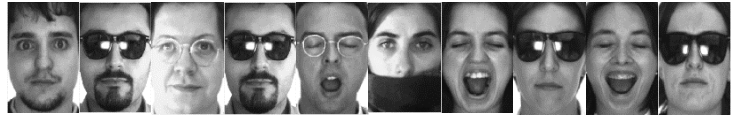

(a)

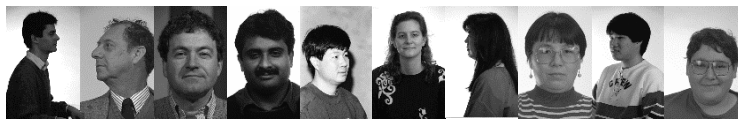

(b)

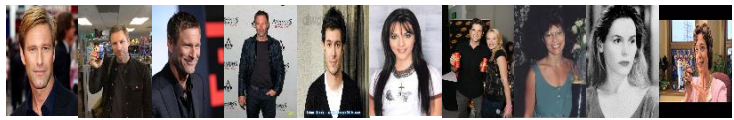

(c)

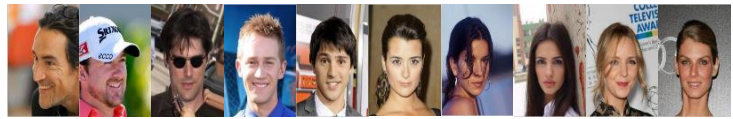

(d)

Fig. 4. Image examples from AR dataset (a), FERET dataset (b), FaceScrub dataset (c) and CeleA dataset (d).

(3) The FaceScrub dataset comprises a total of 107081 images of 530 celebrities, about 200 images each. These images were retrieved from the Internet or taken under real-world situation, with the duplicate and degenerate images removed. From the dataset, 34250 female images and 35718 male images were used in the experiments, where 30169 female images and 31648 male images were selected for training, and the rest for testing.

(4) The CelebA dataset is composed of 118162 female images and 84437 male images, covering large pose variations and background clutter. From the dataset, 80000 female images and 80000 male images were selected for training. From the rest, 4000 male images and 4000 female images were chosen for testing.

It should be noted that for each of these four datasets, all images for any single subject are either in the training set or the testing set, but not both. We describe the standard CNN in Table 1 and report the results in Tables 2-5, bolding the highest accuracies. Note that SI=00000 means the standard CNN.

In Table 2, all S-CNNs outperform the CNN (92.30%) in terms of accuracy. With only one shortcut CPL, i.e. the shortcut styles of 10000, 01000, 00100, 00010 and 00001, S-CNNs gradually have slightly worse performance. With two shortcut CPLs, the highest accuracy of S-CNNs is 95.23% obtained by the 01010 shortcut style, and the lowest is 93.65% by the 00011 shortcut style. With three shortcut CPLs, the highest accuracy is 94.83% obtained by the 01101 shortcut style, and the lowest is 93.84% by the 11010 shortcut style. With four shortcut CPLs, the highest accuracy is 94.64% obtained by the 01111 shortcut style, and the lowest is 93.85% by the 11011 shortcut style.

Table 1 Description of the standard CNN used for gender classification.

| layer | type | activation function | patch size | stride | output size |
|---|---|---|---|---|---|
| $\mathbf{x}$ | Input | | | | 32×32×1 |
| $\mathbf{h}_1$ | CL | ReLU | 5×5 | 1 | 28×28×6 |
| $\mathbf{h}_2$ | PL | max-pooling | 2×2 | 2 | 14×14×6 |
| $\mathbf{h}_3$ | CL | ReLU | 5×5 | 1 | 10×10×12 |
| $\mathbf{h}_4$ | PL | max-pooling | 2×2 | 2 | 5×5×12 |
| $\mathbf{h}_5$ | CL | ReLU | 2×2 | 1 | 4×4×16 |
| $\mathbf{h}_6$ | PL | max-pooling | 2×2 | 2 | 2×2×16 |
| $\mathbf{h}_7$ | FCL | | | | 64 |
| $\mathbf{o}$ | Output | softmax | | | 2 |

Table 2 Test accuracies (%) on the AR dataset.

| SI | Accuracy | SI | Accuracy | SI | Accuracy | SI | Accuracy |
|---|---|---|---|---|---|---|---|
| 00000 | 92.30 | 10010 | 93.85 | 11100 | 94.05 | 01011 | 94.45 |
| 10000 | 94.26 | 10001 | 93.84 | 11010 | 93.84 | 00111 | 94.24 |
| 01000 | 94.25 | 01100 | 94.46 | 11001 | 94.06 | 11110 | 94.62 |
| 00100 | 94.21 | 01010 | **95.23** | 10110 | 93.85 | 11101 | 94.42 |
| 00010 | 93.70 | 01001 | 95.22 | 10101 | 94.06 | 11011 | 93.85 |
| 00001 | 93.47 | 00110 | 94.44 | 10011 | 94.05 | 10111 | 94.61 |
| 11000 | 93.68 | 00101 | 93.86 | 01110 | 94.83 | 01111 | 94.64 |
| 10100 | 93.86 | 00011 | 93.65 | 01101 | 94.42 | 11111 | 94.44 |

In Table 3, all S-CNNs but the three shortcut styles of 11100, 11110 and 11111 perform better than the CNN (85.29%). With one shortcut CPL, the highest accuracy of S-CNNs is 89.07% achieved by the shortcut style of 01000, and the lowest is 86.75% by 00001. With two shortcut CPLs, the highest accuracy is 88.55% achieved by the shortcut style of 01001, and the lowest is 85.36% by 00011. With three shortcut CPLs, the highest accuracy is 88.51% achieved by the shortcut style of 10110, and the lowest is 85.17% by 11100. With four shortcut CPLs, the highest accuracy is **89.81**% achieved by the shortcut style of 10111, and the lowest is 83.73% by 11110.

Table 3 Test accuracies (%) on the FERET dataset.

| SI | Accuracy | SI | Accuracy | SI | Accuracy | SI | Accuracy |
|---|---|---|---|---|---|---|---|
| 00000 | 85.29 | 10010 | 87.35 | 11100 | 85.17 | 01011 | 86.96 |
| 10000 | 89.05 | 10001 | 86.44 | 11010 | 85.72 | 00111 | 85.50 |
| 01000 | 89.07 | 01100 | 87.01 | 11001 | 88.25 | 11110 | 83.73 |
| 00100 | 88.74 | 01010 | 87.15 | 10110 | 88.51 | 11101 | 88.80 |
| 00010 | 88.29 | 01001 | 88.55 | 10101 | 86.47 | 11011 | 89.68 |
| 00001 | 86.75 | 00110 | 86.95 | 10011 | 86.16 | 10111 | **89.81** |
| 11000 | 88.26 | 00101 | 86.61 | 01110 | 86.28 | 01111 | 87.92 |
| 10100 | 87.50 | 00011 | 85.36 | 01101 | 86.31 | 11111 | 82.62 |

In Table 4, all S-CNNs have higher accuracies than the CNN (78.57%). With one shortcut CPL, the highest accuracy of S-CNNs is 80.98% reached by the style of 00100, and the lowest is 79.98% by 10000. With two shortcut CPLs, the highest accuracy is **82.14**% reached by the style of 01001, and the lowest is 80.01% by 11000. With three shortcut CPLs, the highest accuracy is 81.37% reached by 11010, and the lowest is 80.52% by 10011 and 00111. With four shortcut CPLs, the highest accuracy is 81.58% reached by 10111, and the lowest is 79.79% by 11011.

Table 4 Test accuracies (%) on the FaceScrub dataset.

| SI | Accuracy | SI | Accuracy | SI | Accuracy | SI | Accuracy |
|---|---|---|---|---|---|---|---|
| 00000 | 78.57 | 10010 | 80.04 | 11100 | 80.55 | 01011 | 80.67 |
| 10000 | 79.98 | 10001 | 80.53 | 11010 | 81.37 | 00111 | 80.52 |
| 01000 | 80.68 | 01100 | 80.96 | 11001 | 80.79 | 11110 | 80.72 |
| 00100 | 80.98 | 01010 | 81.10 | 10110 | 81.07 | 11101 | 80.56 |
| 00010 | 80.6 | 01001 | **82.14** | 10101 | 80.57 | 11011 | 79.79 |
| 00001 | 80.97 | 00110 | 80.85 | 10011 | 80.52 | 10111 | 81.58 |
| 11000 | 80.01 | 00101 | 80.99 | 01110 | 81.21 | 01111 | 80.80 |
| 10100 | 80.37 | 00011 | 80.93 | 01101 | 80.74 | 11111 | 80.59 |

Table 5 Test accuracies (%) on CelebA dataset.

| SI | Accuracy | SI | Accuracy | SI | Accuracy | SI | Accuracy |
|---|---|---|---|---|---|---|---|
| 00000 | 84.21 | 10010 | 86.62 | 11100 | 86.73 | 01011 | 86.39 |
| 10000 | 86.30 | 10001 | 86.20 | 11010 | **87.19** | 00111 | 86.19 |
| 01000 | 85.95 | 01100 | 86.62 | 11001 | 86.15 | 11110 | 86.73 |
| 00100 | 85.92 | 01010 | 86.57 | 10110 | 87.06 | 11101 | 86.67 |
| 00010 | 85.81 | 01001 | 86.29 | 10101 | 86.74 | 11011 | 86.64 |
| 00001 | 85.72 | 00110 | 86.26 | 10011 | 86.63 | 10111 | 86.62 |
| 11000 | 86.18 | 00101 | 86.17 | 01110 | 86.54 | 01111 | 86.75 |
| 10100 | 86.89 | 00011 | 86.54 | 01101 | 86.40 | 11111 | 87.00 |

In Table 5, all S-CNNs have accuracies exceeding the CNN (84.21%). With one shortcut CPL, S-CNNs perform slightly worse gradually for the styles of 10000, 01000, 00100, 00010 and 00001. With two shortcut CPLs, the highest accuracy of S-CNNs is 86.89% attained by 10100, and the lowest is 86.17% by 00101. With three shortcut CPLs, the highest accuracy is **87.19**% attained by 11010, and the lowest is 86.15% by 11001. With four shortcut CPLs, the highest accuracy is 86.75% attained by 01111, and the lowest is 86.62% by 10111.

Overall, the S-CNNs get the highest accuracies of 95.23% on AR, 89.81% on FERET, 82.14% on FaceScrub, and 87.19% on CeleA. Compared to the CNN, these accuracies gain a relative increase of 3.17%, 5.30%, 4.54%, and 3.54%, respectively. This is probably because the S-CNNs can integrate multi-scale features from many CPLs, leading to a more suitable model. It should be noted that the best shortcut styles are generally data-dependent, varying on different datasets. Furthermore, in this experiment the shortcut style of 11111 is never the best one for S-CNNs, probably with too many parameters to get well-trained.

4.2 TEXTURE CLASSIFICATION

In this subsection, using the same CNN architecture except with the output size of 61 given in Table 1, we investigate performance of the CNN and S-CNNs on CUReT dataset (Dana et al, 1999) for texture classification. The CUReT dataset has 12505 images in 61 texture classes, 205 images per class, with different pose and illumination conditions, specularities, shadowing, and surface normal variations. From the dataset, 185 images per class were selected for training, and the rest for testing. Examples of this dataset are shown in Fig. 5. For 60000 training iterations, the results of the CNN and S-CNNs are reported in Table 6, with the highest accuracy bolded.

In Table 6, all S-CNNs have higher accuracies than the CNN (66.17%). For 1-4 shortcut CPLs, the highest accuracies of S-CNNs are 75.12%, 78.86%, **79.00**% and 75.81% with the lowest of 66.97%, 64.87%, 70.39% and 73.00%, respectively. The highest ones are achieved by the shortcut styles of 00001, 00011, 01101 and 11101, and the lowest by 10000, 11000, 10110 and 01111.

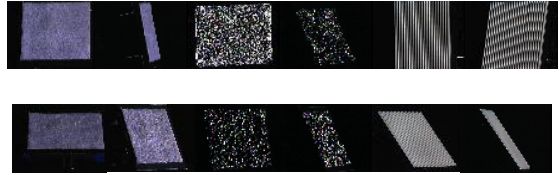

Fig. 5. Image examples from the CUReT dataset.

Table 6 Test accuracies (%) on the CUReT dataset.

| SI | Accuracy | SI | Accuracy | SI | Accuracy | SI | Accuracy |
|---|---|---|---|---|---|---|---|
| 00000 | 66.17 | 10010 | 69.95 | 11100 | 73.42 | 01011 | 77.35 |
| 10000 | 66.97 | 10001 | 72.44 | 11010 | 73.58 | 00111 | 77.24 |
| 01000 | 71.70 | 01100 | 76.86 | 11001 | 70.39 | 11110 | 74.00 |
| 00100 | 73.42 | 01010 | 74.38 | 10110 | 71.87 | 11101 | 75.81 |
| 00010 | 74.91 | 01001 | 70.24 | 10101 | 74.69 | 11011 | 75.14 |
| 00001 | 75.12 | 00110 | 75.97 | 10011 | 75.49 | 10111 | 75.38 |
| 11000 | 64.87 | 00101 | 77.34 | 01110 | 76.13 | 01111 | 73.00 |
| 10100 | 72.13 | 00011 | 78.86 | 01101 | **79.00** | 11111 | 74.72 |

Overall, the highest accuracy of S-CNNs on CUReT is 79.00%, relatively increased by 19.39% in comparison with the CNN. This means that the S-CNNs could be more suitable than the CNN for texture classification by integration of multi-scale features. However, in general the best shortcut style is not 11111, by which the S-CNN gets the accuracy of 74.72%.

4.3 DIGIT RECOGNITION

Taking the LeNet as the CNN (LeCun et al, 1998), we now move forward to test the CNN and S-CNNs on MNIST dataset for digit recognition. The MNIST dataset consists of 28×28 pixel grayscale images of hand-written digits (from 0 to 9) (Yu et al, 2013). There are 60000 training images and 10000 testing images in total, but noting that the number of images per digit is not uniformly distributed. We use the standard split for training and testing here.

The LeNet contains two convolutional layers and two pooling layers with the fully-connected layer being the vectorization of the last pooling layer. For 10000 training iterations, the results of the CNN and S-CNNs are reported in Table 7.

Table 7 Test accuracies (%) on the MNIST dataset.

| SI | Accuracy | SI | Accuracy |
|---|---|---|---|
| 000 | 99.04 | 110 | 99.04 |
| 100 | 99.06 | 101 | 99.08 |
| 010 | **99.20** | 011 | 99.17 |
| 001 | 99.16 | 111 | 99.13 |

From Table 7, we can see that the S-CNNs outperform the CNN (99.04%) overall, in consistency with that for gender and texture classification. The highest accuracy is **99.20**% by the shortcut style of 010 for one shortcut CLP, and 99.17% by 110 for two shortcut CPLs, with the lowest accuracies of 99.06% and 99.04% by 100 and 110, respectively.

Overall, the S-CNNs get the highest accuracy of 99.20% on the MNIST dataset, gaining a relative 0.16% increase compared with the CNN. Note that the best performance is obtained by the style of 010, rather than by 111.

Table 8 The CNN architecture used for the CIFAR-10 dataset.

| layer | type | activation function | patch size | stride | output size |
|---|---|---|---|---|---|
| $x$ | Input | | | | 32×32×3 |
| $h_1$ | CL | ReLU | 5×5 | 1 | 32×32×32 |
| $h_2$ | PL | max-pooling | 3×3 | 2 | 16×16×32 |
| $h_3$ | CL | ReLU | 5×5 | 1 | 16×16×32 |
| $h_4$ | PL | max-pooling | 3×3 | 2 | 8×8×32 |
| $h_5$ | CL | ReLU | 2×2 | 1 | 8×8×32 |
| $h_6$ | PL | max-pooling | 3×3 | 2 | 4×4×16 |
| $h_7$ | FCL | | | | 256 |
| $o$ | Output | Softmax | | | 10 |

Table 9 Test accuracies (%) on the CIFAR-10 dataset.

| SI | Accuracy | SI | Accuracy |
|---|---|---|---|
| 00000 | 73.10 | 00001 | **77.57** |
| 10000 | 23.02 | 00101 | 70.00 |
| 01000 | 40.64 | 00011 | 74.37 |
| 00100 | 74.61 | 00111 | 69.64 |

4.4 OBJECT RECOGNITION

Using the same architecture described in Table 8 except with each pooling layer modified by the local response normalization (LRN) function (Krizhevsky et al, 2012), we evaluate performance of the CNN and S-CNNs on the CIFAR-10 dataset (Krizhevsky. 2012) for object recognition. The CIFAR-10 dataset is a set of color images of 32×32 pixels. It contains 60000 images of 10 commonly seen object categories (e.g., animals and vehicles), varying significantly not only in object position and object scale within each class but also in colors and textures among classes. There are 50000 images used for training and the rest 10000 for testing, and all 10 categories have equal number of training and test images. We use the standard split for training and testing.

For 60000 training iterations, we report the results of the CNN and seven S-CNNs in Table 9. We do not give the results of all short-cut styles because they generally deteriorate in case of concatenating too many CPLs. Even so, the best S-CNN can achieve the highest accuracy of **77.57**% on the CIFAR-10 dataset, gaining a relative increase of 6.11% compared with the CNN. Thus, a proper shortcut style is important to make better performance of S-CNNs.

4.5 DIFFERENT SETTINGS

Using the same CNN architecture described in Table 1, we further compare the CNN and S-CNNs with different settings of pooling schemes, activation functions, initializations, and optimizations on the AR dataset. The results are reported in Tables 10-13, with the best accuracies bolded.

In Table 10, we show performance of the CNN and S-CNNs with average pooling (LeCun et al, 1990) instead of max-pooling. Compared to 92.30% in Table 2, the CNN has a relative reduction of 4.12% in accuracy, whereas to 95.23%, the highest accuracy of S-CNNs is 94.85%, relatively reduced by 0.40%.

Table 10 Test accuracies (%) with average pooling on the AR dataset.

| SI | Accuracy | SI | Accuracy | SI | Accuracy | SI | Accuracy |
| --- | --- | --- | --- | --- | --- | --- | --- |
| 00000 | 88.50 | 10010 | 93.66 | 11100 | 93.27 | 01011 | 93.68 |
| 10000 | 93.66 | 10001 | 93.67 | 11010 | 93.44 | 00111 | 94.45 |
| 01000 | 93.48 | 01100 | 93.13 | 11001 | 93.48 | 11110 | 93.86 |
| 00100 | 92.90 | 01010 | 94.42 | 10110 | 93.67 | 11101 | 93.85 |
| 00010 | 92.88 | 01001 | 94.45 | 10101 | 93.67 | 11011 | 92.73 |
| 00001 | 92.12 | 00110 | 94.46 | 10011 | 94.45 | 10111 | 93.28 |
| 11000 | 93.08 | 00101 | 93.48 | 01110 | **94.85** | 01111 | 94.24 |
| 10100 | 93.27 | 00011 | 92.13 | 01101 | 93.31 | 11111 | 94.05 |

In Table 11, we describe their performance with ReLU replaced by sigmoid. With a relative reduction of 26.68%, the accuracy of the CNN decreases from 92.30% to 67.67%, whereas the highest accuracy of S-CNNs drops from 95.23% to 94.85%, relatively reduced by 0.40%.

Table 11 Test accuracies (%) with sigmoid function on the AR dataset.

| SI | Accuracy | SI | Accuracy | SI | Accuracy | SI | Accuracy |
| --- | --- | --- | --- | --- | --- | --- | --- |
| 00000 | 67.67 | 10010 | 93.65 | 11100 | 94.45 | 01011 | 91.34 |
| 10000 | 93.68 | 10001 | 93.64 | 11010 | 94.41 | 00111 | 85.05 |
| 01000 | 93.00 | 01100 | 94.06 | 11001 | 94.22 | 11110 | **94.85** |
| 00100 | 88.10 | 01010 | 94.00 | 10110 | 93.67 | 11101 | 94.84 |
| 00010 | 67.93 | 01001 | 92.00 | 10101 | 93.70 | 11011 | 94.27 |
| 00001 | 67.67 | 00110 | 85.60 | 10011 | 93.70 | 10111 | 94.25 |
| 11000 | 94.26 | 00101 | 90.23 | 01110 | 94.00 | 01111 | 90.29 |
| 10100 | 93.70 | 00011 | 69.29 | 01101 | 91.78 | 11111 | 94.06 |

In Table 12, we depict their performance with a different initialization of Msra (He et al, 2015) from Xavier. It can be seen that the CNN has a relative 0.88% reduction, from 92.30% to 91.49%. However, S-CNNs has a relative 0.64% reduction in terms of the highest accuracy, from 95.23% to 94.62%.

Table 12 Test accuracies (%) with Msra initialization on the AR dataset.

| SI | Accuracy | SI | Accuracy | SI | Accuracy | SI | Accuracy |
| --- | --- | --- | --- | --- | --- | --- | --- |
| 00000 | 91.49 | 10010 | 94.35 | 11100 | 94.25 | 01011 | 92.32 |
| 10000 | 93.47 | 10001 | 94.24 | 11010 | 94.21 | 00111 | 93.49 |
| 01000 | 94.04 | 01100 | 94.24 | 11001 | 93.86 | 11110 | 94.06 |
| 00100 | 92.53 | 01010 | 94.24 | 10110 | 93.73 | 11101 | 94.05 |
| 00010 | 91.97 | 01001 | 92.70 | 10101 | 93.66 | 11011 | 93.86 |
| 00001 | 91.90 | 00110 | 92.52 | 10011 | 93.56 | 10111 | 92.90 |
| 11000 | 93.67 | 00101 | 93.69 | 01110 | 94.02 | 01111 | 92.13 |
| 10100 | **94.62** | 00011 | 93.49 | 01101 | 92.91 | 11111 | 93.85 |

In Table 13, we delineate their performance with the Adam (Kingma and Ba, 2015) optimization. We can clearly see that, the CNN achieves the accuracy of 91.35%, and the best S-CNN 94.84%. They have a relative reduction of 1.03% and 0.41% in comparison with 92.30% and 95.23%, respectively.

Table 13 Test accuracies (%) with Adam optimization on the AR dataset

| SI | Accuracy | SI | Accuracy | SI | Accuracy | SI | Accuracy |
|---|---|---|---|---|---|---|---|
| 00000 | 91.35 | 10010 | 94.23 | 11100 | 94.04 | 01011 | **94.84** |
| 10000 | 93.37 | 10001 | 93.66 | 11010 | 93.65 | 00111 | 92.72 |
| 01000 | 93.88 | 01100 | 93.47 | 11001 | 93.09 | 11110 | 94.80 |
| 00100 | 93.47 | 01010 | 93.09 | 10110 | 93.09 | 11101 | 94.43 |
| 00010 | 93.45 | 01001 | 92.34 | 10101 | 93.03 | 11011 | 94.04 |
| 00001 | 93.27 | 00110 | 94.06 | 10011 | 92.89 | 10111 | 92.70 |
| 11000 | 93.85 | 00101 | 93.86 | 01110 | 94.07 | 01111 | 92.33 |
| 10100 | 94.23 | 00011 | 91.53 | 01101 | 93.86 | 11111 | 92.52 |

Overall, on the AR dataset the S-CNNs have a smaller accuracy reduction than the CNN with different settings, especially of the sigmoid function. This indicates that the S-CNNs have more stable performance than the CNN in general.

Table 14 Test accuracies (%) with different number of convolutional kernels on the CUReT dataset.

| SI | Accuracy | SI | Accuracy | SI | Accuracy | SI | Accuracy |
|---|---|---|---|---|---|---|---|
| 00000 | 61.71 | 10010 | 68.02 | 11100 | 72.33 | 01011 | 77.89 |
| 10000 | 62.52 | 10001 | 68.65 | 11010 | 66.81 | 00111 | 72.34 |
| 01000 | 73.69 | 01100 | 75.22 | 11001 | 68.32 | 11110 | 72.35 |
| 00100 | 73.70 | 01010 | 71.99 | 10110 | 73.87 | 11101 | 72.92 |
| 00010 | 73.72 | 01001 | 70.22 | 10101 | 75.07 | 11011 | 74.88 |
| 00001 | 68.11 | 00110 | 75.05 | 10011 | 74.13 | 10111 | 74.88 |
| 11000 | 62.37 | 00101 | 74.61 | 01110 | 74.94 | 01111 | 74.95 |
| 10100 | 67.94 | 00011 | 74.69 | 01101 | **78.73** | 11111 | 72.90 |

## 4.6 DIFFERENT NUMBERS AND SIZES OF CONVOLUTONAL KERNELS

Finally, we examine performance of the CNN and S-CNNs with different numbers and sizes of convolutional kernels on the CUReT dataset.

In Table 14, we report the results using the same architecture except with 10 kernels for each of the three convolutional layers in Table 1. It can be seen that, the CNN gets the accuracy of 61.71%, and the best S-CNN 78.73%. They have a relative accuracy reduction of 6.74% and 0.34%, respectively compared to 66.17% and 79.00% in Table 6.

Table 15 Test accuracies (%) of the CNN and S-CNNs with different size of convolutional kernels on the CUReT dataset

| SI | Accuracy | SI | Accuracy | SI | Accuracy | SI | Accuracy |
|---|---|---|---|---|---|---|---|
| 00000 | 64.97 | 10010 | 72.34 | 11100 | 72.44 | 01011 | 76.06 |
| 10000 | 67.02 | 10001 | 72.76 | 11010 | 68.37 | 00111 | **78.77** |
| 01000 | 71.92 | 01100 | 74.28 | 11001 | 71.11 | 11110 | 72.00 |
| 00100 | 71.34 | 01010 | 78.05 | 10110 | 71.47 | 11101 | 71.90 |
| 00010 | 69.89 | 01001 | 75.25 | 10101 | 73.59 | 11011 | 69.53 |
| 00001 | 70.52 | 00110 | 76.41 | 10011 | 75.96 | 10111 | 73.24 |
| 11000 | 70.78 | 00101 | 77.71 | 01110 | 73.02 | 01111 | 76.54 |
| 10100 | 72.20 | 00011 | 76.80 | 01101 | 77.08 | 11111 | 75.08 |

In Table 15, we report the results using the same architecture except with kernel sizes of $7 \times 7$,

$4\times4$ and $2\times2$ for the three convolutional layers in Table 1. It can be seen that, the CNN gets the accuracy of 64.97%, and the best S-CNN 78.77%. They have a relative accuracy reduction of 1.81% and 0.29%, respectively.

Therefore, with different number and size of convolutional kernels, the S-CNNs have a relatively smaller reduction of performance than the CNN on the CUReT dataset, indicating their less insensitiveness.

## 5 CONCLUSIONS

In this paper, we have presented a concatenating framework of S-CNNs, which can integrate multi-scale features through shortcut connections in a CNN. Also, we compare performance of the CNN and S-CNNs on four different tasks, including gender classification, texture classification, digit recognition, and object recognition. Based on extensive experiments, we show that the S-CNNs can produce higher accuracies than the CNN in these tasks, especially in texture classification and gender classification. Moreover, the S-CNNs have more stable performance than the CNN with different settings of pooling schemes, activation functions, initializations, and optimizations. Additionally, the S-CNNs are less insensitive to kernel numbers and kernel sizes. Therefore, we conclude that the S-CNNs can improve performance of the CNN by integrating multi-scale features in the proposed concatenating framework, although the S-CNNs may have a performance reduction in case of concatenating too many features, which are likely to contain much redundant information and even to make training very difficult.

It should be noted that the best shortcut style is dataset-dependent. Different datasets may have different best shortcut styles. The shortcut style that all hidden layers are concatenated to the fully-connected layer cannot always guarantee the best performance, which was advocated in DenseNets (Huang et al, 2017). As future work, we will study the problem of how to dynamically determine the best shortcut style for a dataset in theory and practice.

## ACKNOELWDGMENTS

This work was supported in part by the National Natural Science Foundation of China under Grant 61175004, the Natural Science Foundation of Beijing under grant 4112009, the Specialized Research Fund for the Doctoral Program of Higher Education of China under Grant 20121102110029, the China Postdoctoral Science Foundation funded project under Grant 2015M580952, and the project supported by Beijing Postdoctoral Research Foundation under Grant 2016ZZ-24.